\pdfoutput=1

\documentclass[11pt]{article}

\usepackage{emnlp2021}

\usepackage{times}
\usepackage{latexsym}

\usepackage[T1]{fontenc}

\usepackage[utf8]{inputenc}

\usepackage{times}
\usepackage{latexsym}

\usepackage{url}
\usepackage{times}
\usepackage{amssymb} 
\usepackage{microtype}
\usepackage{graphicx}
\usepackage{booktabs}
\usepackage{floatrow}
\usepackage{booktabs}
\usepackage{numprint}
\usepackage{subcaption} 
\usepackage{enumerate}
\usepackage{todonotes}
\usepackage{bm}
\usepackage{multirow}
\usepackage{arydshln}
\usepackage{xcolor,soul}
\usepackage{amssymb}
\usepackage{pifont}
\usepackage{xurl}

\sethlcolor{lightgray}

\newcommand{\laurent}[1]{\textcolor{black}{ #1}}

\newcommand{\ahmet}[1]{\textcolor{black}{ #1}}
\newcommand{\alex}[1]{\textcolor{black}{ #1}}

\DeclareUnicodeCharacter{0300}{\`o}
\DeclareUnicodeCharacter{01A1}{\^e}
\DeclareUnicodeCharacter{01B0}{\^o}

\newcommand{\mbart}{\textsc{Mbart-ft}~}
\newcommand{\task}{\textsc{Task Adapters}~}

\newcommand{\deo}{\textsc{Denoising Adapters}}

\title{Multilingual Unsupervised Neural Machine Translation \\
with Denoising Adapters}

\usepackage{enumitem}

\setlist[itemize]{leftmargin=*}

\author{Ahmet \"{U}st\"{u}n\textsuperscript{\textdagger}\thanks{~~Work done during an
    internship at NAVER LABS Europe.} 
\qquad Alexandre Bérard\textsuperscript{\textdaggerdbl} 
\qquad Laurent Besacier\textsuperscript{\textdaggerdbl} 
\qquad Matthias Gallé\textsuperscript{\textdaggerdbl}  
\vspace{.2cm} \\ 
\textsuperscript{\textdagger}University of Groningen \qquad 
\textsuperscript{\textdaggerdbl}NAVER LABS Europe \vspace{.1cm} \\
\tt{a.ustun@rug.nl} \\
\tt{first.last@naverlabs.com}
}

\date{}

\begin{document}
\maketitle
\begin{abstract}
We consider the problem of multilingual unsupervised machine translation, translating to and from languages that only have monolingual data by using auxiliary parallel language pairs.
For this problem the standard procedure so far to leverage the monolingual data is \textit{back-translation}, 
which is computationally costly and hard to tune.

In this paper we propose instead to use \textit{denoising adapters}, adapter layers with a denoising objective, on top of pre-trained mBART-50. 
In addition to the modularity and flexibility of such an approach we show that the resulting translations are on-par with back-translating as measured by BLEU, and furthermore it allows adding unseen languages incrementally.
\end{abstract}

\section{Introduction}
\label{intro}

Two major trends have in the last years provided surprising and exciting new avenues in Neural Machine Translation (NMT).
First, Multilingual Neural Machine Translation \cite{firat2016multi,ha2016toward,johnson2017google,aharoni2019massively} has achieved 
impressive results on large-scale multilingual benchmarks with diverse sets of language pairs. 
It has the advantage of resulting in only one model to maintain, as well as benefiting from cross-lingual knowledge transfer.
Second, Unsupervised Neural Machine Translation (UNMT) \cite{lample2018unsupervised,artetxe2018unsupervised} allows to train translation systems from monolingual data only.
Training bilingual UNMT systems \cite{lample2019cross,artetxe-etal-2019-effective} often assume high-quality in-domain monolingual data and is mostly limited to resource-rich languages. 
In addition to the pretraining and the denoising auto-encoding, they require one or more expensive steps of back-translation \cite{sennrich-etal-2016-improving} in order to create an artificial  parallel training corpus.

\begin{figure}[t]
    \includegraphics[scale=0.5]{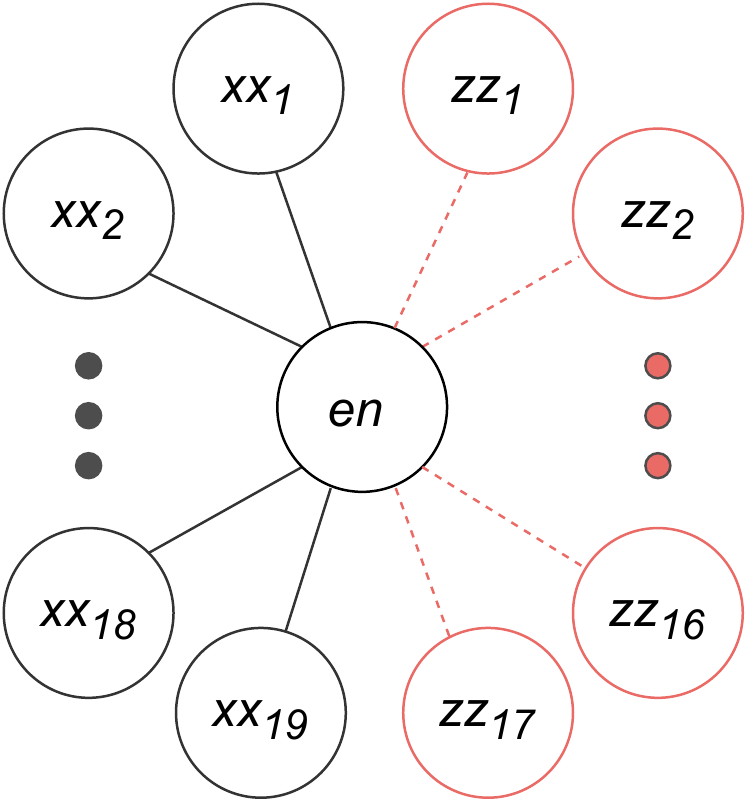}
    \caption{Overview of our multilingual unsupervised NMT setup where dashed lines indicate 17 unsupervised languages without parallel data ($\textrm{zz}_n$) and full lines indicate 19 auxiliary languages with parallel data for training ($\textrm{xx}_n$). Adapted from~\citet{garcia2020harnessing}.}
  \label{fig:mUNMT}
\end{figure}

Multilingual UNMT aims at combining these two trends. As depicted in Fig~\ref{fig:mUNMT}, some \emph{auxiliary} languages have access to parallel data paired with English ($en\leftrightarrow xx_1$), while \emph{unsupervised} languages only have monolingual data ($zz_1$). The goal of such an approach is to make use of the auxiliary parallel data to learn the translation task and hopefully transfer this task knowledge to the unsupervised languages. The end model should be able to translate to/from English in both the auxiliary and unsupervised languages.

This setting has only been addressed very recently~\citep{sun-etal-2020-knowledge-distillation,liu-etal-2020-multilingual,wang2020cross,garcia2020harnessing}. However all current approaches rely on back-translation, either \textit{offline} or \textit{online}.
This is computationally costly and it requires a lot of engineering effort when applied to large-scale setups. 

In this paper, we propose a 2-step approach based on \textbf{denoising adapters} that enable modular multilingual unsupervised NMT \textbf{without back-translation}. Our approach combines monolingual denoising adapters with multilingual transfer learning on auxiliary parallel data. More precisely our denoising adapters are lightweight adapter modules inserted into multilingual BART \cite[mBART]{liu-etal-2020-multilingual} and trained with a \textit{denoising} objective on monolingual data for each language separately. The first step, i.e. monolingual training, allows learning of language-specific encoding and decoding through adapter modules which can easily be composed with other languages' adapters for translation. The second step transfers mBART to multilingual UNMT by plugging in our denoising adapters and then fine-tuning cross-attention with auxiliary parallel data. Our approach also allows extending mBART with new languages which are not included in pretraining as shown in Sect.~\ref{sec:new-languages}. This means that denoising adapters can be trained \textbf{incrementally} after mBART fine-tuning to add \textit{any} new language to the existing setup.

In our experiments, we train denoising adapters for 17 diverse unsupervised languages together with 20 auxiliary languages and evaluate the final model on TED talks \cite{qi-etal-2018-pre}. Our results show that our approach is on par with back-translation for a majority of languages while being more modular and efficient. Moreover, using denoising adapters jointly with back-translation further improves unsupervised translation performance.  

\paragraph{Contributions}

In summary, we make the following contributions: 
\textbf{1)} We propose denoising adapters, monolingually-trained adapter layers to leverage monolingual data for unsupervised machine translation. 
\textbf{2)} We introduce a 2-step approach for multilingual UNMT using denoising adapters and multilingual fine-tuning of mBART's cross-attention with auxiliary parallel data.
\textbf{3)} We conduct experiments on a large set of language pairs showing effectiveness of denoising adapters with and without back-translation. 
\textbf{4)} Finally, we provide further analysis to the use of denoising adapters such as extending mBART with completely new languages.

\section{Background}

\subsection{mBART fine-tuning for translation}

Multilingual BART, mBART \cite{liu-etal-2020-multilingual}, is a Transformer-based sequence-to-sequence model that consists of an encoder and an autoregressive decoder (hence \underline{B}idirectional and \underline{A}uto-\underline{R}egressive \underline{T}ransformer). 
It is pretrained by reconstructing, i.e.~\textit{denoising} the original text from a noisy version corrupted with a set of noising functions. 
Although in the original BART \cite{lewis-etal-2020-bart}, several noising functions were introduced such as token masking, token deletion, word-span masking, sentence permutation and document rotation; mBART uses only text infilling (which is based on span masking) and sentence permutation. 
Architecture-wise, mBART is a Transformer model~\citep{vaswani2017attention} with 12 encoder and 12 decoder layers with hidden dimension of 1024 and 16 attention heads. It has a large multilingual vocabulary of 250k tokens obtained from 100 languages.
To fine-tune mBART to machine translation, the weights of the pretrained model are loaded and \textit{all} parameters are trained with parallel data either in a bilingual \cite{liu-etal-2020-multilingual} or a multilingual setup \cite{cooper-stickland-etal-2021-recipes, tang2020multilingual} to leverage the full capacity of multilingual pretraining.

In our experiments we use mBART-50\footnote{To simplify notation we will refer to mBART-50 as mBART} \cite{tang2020multilingual}, which is pretrained on 50 different languages, as both the \textit{parent} model for our adapters and as a strong baseline for multilingual MT fine-tuning. 

\subsection{Adapters for MT}

Adapter modules \cite{houlsby2019parameter}, or simply adapters, are designed to adapt a large pretrained model to a downstream task with lightweight residual layers \cite{rebuffi2018efficient} that are inserted into each layer of the model. The adapter layers are trained on the downstream task's data while keeping the parameters of the original pretrained model (the \textit{parent} model) frozen. This allows a high degree of parameter sharing and avoids \textit{catastrophic forgetting} of the knowledge learned during pretraining. Adapters have mainly been used for parameter-efficient fine-tuning \cite{houlsby2019parameter,stickland2019bert} but they have also been used to learn language-specific information within a multilingual pretrained model in zero-shot settings \cite{ustun-etal-2020-udapter}. Similar to our work, \citet{pfeiffer2020mad} have proposed to learn \textit{language} and \textit{task} adapters via masked language modelling and target task objective respectively to combine them for cross-lingual transfer. However, unlike our approach, they trained adapters for transfer learning from one language to another 
but not in a multilingual setup. Moreover, they focus on sequence classification tasks, which highly differ from sequence-to-sequence tasks such as MT. Our work instead proposes a fully multilingual transfer learning method for unsupervised MT that requires composition of encoder and decoder adapters. 

In machine translation,
\citet{bapna-firat-2019-simple} proposed \textit{bilingual} adapters for improving a pretrained multilingual MT model or for domain adaptation whereas \citet{philip-etal-2020-monolingual} trained language-specific adapters in a multilingual MT setup with a focus on zero-shot MT performance. Finally, \citet{cooper-stickland-etal-2021-recipes} use language-agnostic task adapters for fine-tuning BART and mBART to bilingual and multilingual MT. However, none of these approaches are directly applicable for unsupervised MT task as they train language or task-specific adapters on parallel data. 

\subsection{Multilingual Unsupervised NMT} 

We define Multilingual UNMT as the problem of learning both from parallel data centered in one language (English) and monolingual data for translating between the centre language and any of the provided languages. Prior work~\citep{sen-etal-2019-multilingual,sun-etal-2020-knowledge-distillation} trained a single shared model for multiple language pairs by using a denoising auto-encoder and back-translation. 
\citet{sun-etal-2020-knowledge-distillation} also proposed to use knowledge distillation to enhance multilingual unsupervised translation. 
Another line of research \cite{wang2020cross, li-etal-2020-reference, garcia2020harnessing} has explored the use of auxiliary parallel data in a multilingual UNMT setting. 
These studies employ a standard two-stage training schema \cite{lample2019cross} that consists of a first multi-task pretraining step with denoising and translation objectives, and a second fine-tuning step using back-translation. 
\citet{liu-etal-2020-multilingual} eliminated the back-translation step by fine-tuning the pretrained multilingual model on a language pair (e.g. \textit{hi}$\to$\textit{en}) related to the desired unsupervised language pair (e.g. \textit{ne}$\to$\textit{en}). More similar to our work, \citet{garcia2020harnessing} trained a single model on several unsupervised languages pairs by using monolingual data in those languages plus auxiliary parallel data, following the setup illustrated by Fig.~\ref{fig:mUNMT}.
Furthermore, they leverage synthetic parallel data via offline back-translation \cite{sennrich-etal-2016-improving} and iterative back-translation in subsequent steps to fine-tune their model. 
In contrast to our approach, their method focuses on combining existing back-translation methods with multilingual UNMT in several steps. 
Additionally, their method is based on joint multi-task pretraining for all languages which lacks flexibility for incrementally adding new languages.

\begin{figure}[t]
    \includegraphics[scale=0.6]{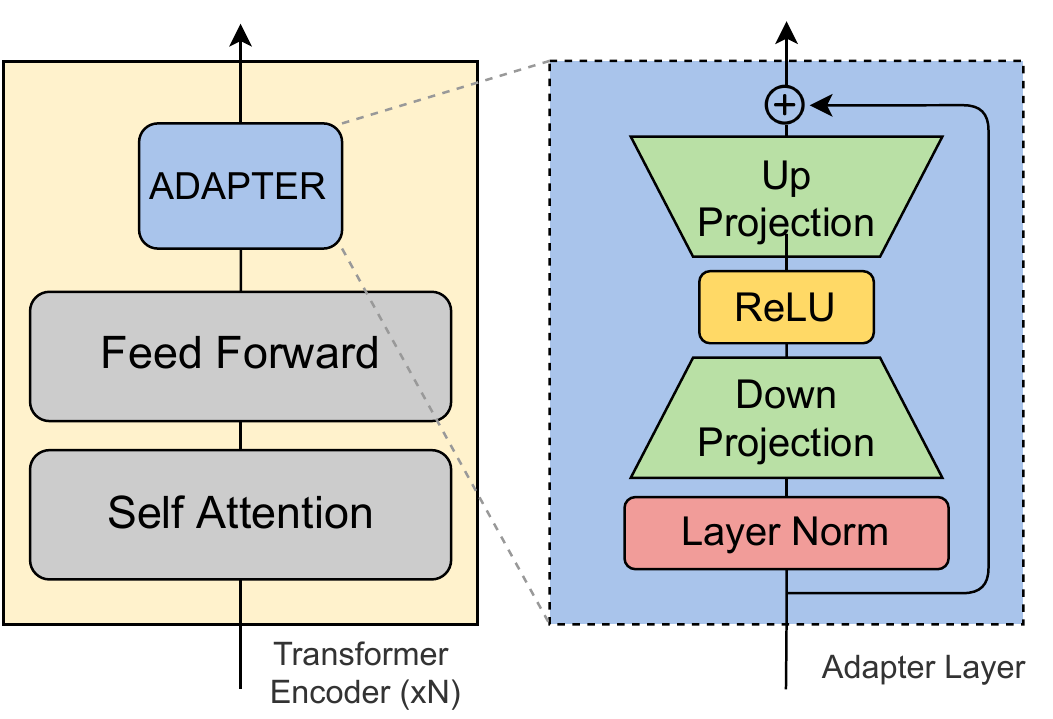}
    \caption{Overview of the adapter architecture that is used in the experiments}
  \label{fig:adapter}
\end{figure}

\begin{figure*}[t]
  \begin{subfigure}[b]{0.46\textwidth}
    \includegraphics[width=\textwidth]{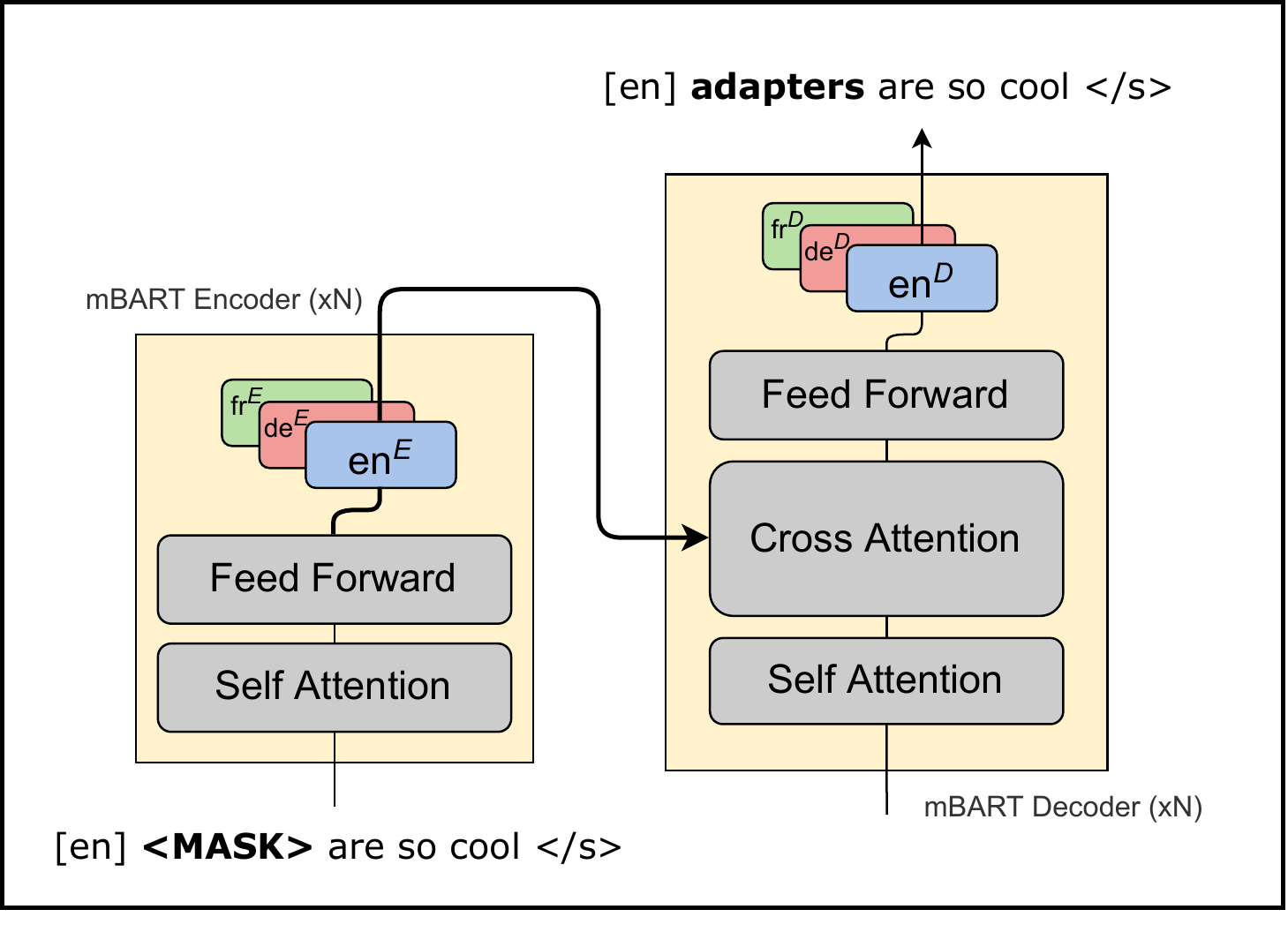}
    \centering
    \caption{\alex{Step 1:} Denoising autoencoding with \textbf{monolingual data}}
    \label{fig:deo}
  \end{subfigure} \hspace{0.8cm}
  \begin{subfigure}[b]{0.46\textwidth}
    \includegraphics[width=\textwidth]{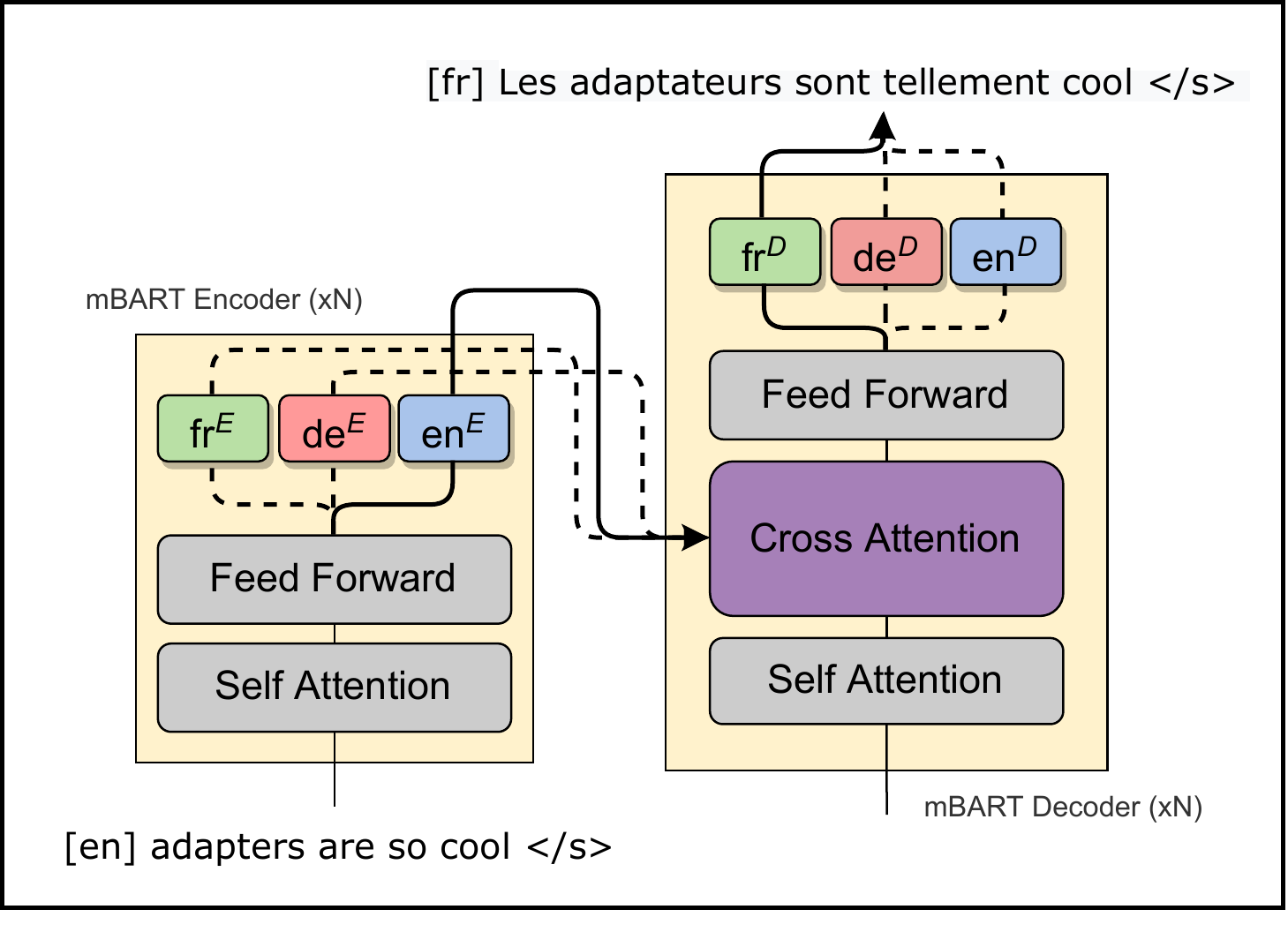}
    \caption{\alex{Step 2:} Multilingual MT training with \textbf{parallel data}}
    \label{fig:nmt}
  \end{subfigure}
  \caption{Overview of \deo{}. In \ref{fig:deo}, denoising adapters (colored boxes) are trained on monolingual data separately for each language, including languages without parallel data. In this step only adapter layers are trained. In \ref{fig:nmt}, \textit{all} denoising adapters that are trained in \ref{fig:deo} are frozen,
  and only the \textit{cross-attention} of mBART \cite{liu-etal-2020-multilingual} is updated with auxiliary parallel data.}
  \label{fig:model}
\end{figure*}

\section{Denoising Adapters for Multilingual Unsupervised MT}
\label{sec:model}

We address the limitations of existing methods mentioned above by proposing denoising adapters for multilingual unsupervised MT. 
Denoising adapters are monolingually-trained language adapters, therefore eliminating the dependence on parallel data.
They allow learning and localizing general-purpose language-specific representations on top of pretrained models such as mBART. 
These denoising adapters can then easily be used for multilingual MT, including unsupervised machine translation without back-translation. 

\paragraph{Architecture}

For our denoising adapters, following \citet{bapna-firat-2019-simple}, we use a simple feed-forward network with a ReLU activation. 
Each adapter module also includes a parametrized normalization layer that acts on the input of the adapter and allows learning the activation pattern of Transformer layers. \ahmet{Figure~\ref{fig:adapter} shows the architecture of an adapter layer.} \ahmet{More formally, a denoising adapter module $\textrm{D}_{i}$ at layer $i$ consists of a layer-normalization LN of the input $z_i\in \mathbb{R}^h$, followed by a down-projection $W_{down}\in \mathbb{R}^{h\times b}$ with bottleneck dimension $b$, a non-linear function and a up projection $W_{up}\in \mathbb{R}^{b\times h}$ combined with a residual connection with the input $z_i$:}
\begin{eqnarray*}
\textrm{D}_{i}(z_i) = W^T_{up} \textrm{ReLU}(W^T_{down} \textrm{LN}(z_i)) + z_i
\end{eqnarray*}
\alex{Bias terms are omitted for clarity. For simplicity, we denote as $\textbf{D}^E=\{\textrm{D}^E_{1\leq{}i\leq{}12}\}$ (resp. $\textbf{D}^D$) the set of encoder (resp. decoder) adapters.}

Similarly to \citet{philip-etal-2020-monolingual}, we insert an adapter module into each layer of the Transformer encoder and decoder, after the feed-forward block; and we train encoder and decoder denoising adapters ($\textbf{D}_{xx}^E$, $\textbf{D}_{xx}^D$) for each language $xx$ in a language-specific manner. This enables to combine encoder adapters $\textbf{D}_{xx}^E$ for source language $xx$ and decoder adapters $\textbf{D}_{yy}^D$ for target language $yy$ to translate from $xx$ to $yy$.

\begin{table*}[ht]
 \small
 \centering
 \begin{tabular}{@{}crrrrrrrrrrrr@{\hskip 0.15in}c@{}}
 \toprule
 \multicolumn{14}{c}{$\bm{zz}\to~\bm{en}$} \\
         & & es & nl & hr & uk & sv & lt & id & fi & et & ur & kk & \textsc{avg-11}   \\ \midrule
 (1) & \textsc{Bilingual} & \textbf{43.4} & \textbf{38.2} & 35.4 & 27.4 & 36.8 & 20.0 & 31.3 & 13.4 & 8.0  & 4.0  & 2.1 & 23.6 \\ \midrule
 & \textsc{Mbart-ft} & 39.8  & 32.8  & 26.7  & 26.6  & 30.0  & 21.0  & 22.6 & 19.7  & 16.8  & 9.2   & 9.6  & 23.2 \\
(2) & \textsc{Task Adapters}   & 42.0  & 35.5  & 32.9  & 30.8  & 38.0  & 25.4 & 33.4 & 22.5  & 21.6  & 20.0  & 12.9  & 28.6 \\
 & \textsc{Denois. Adapters}   & 42.3  & 37.0  & 38.0  & 31.1  & 42.2 & 31.2 & 34.8 & 25.2  & 28.6  & \textbf{24.3}  & 15.6  & 31.8 \\ \midrule
 & \textsc{Mbart-ft (+bt)} & 40.4 & 33.6 & 27.0 & 27.4 & 32.5 & 22.1 & 24.5 & 21.6 & 18.0 & 6.6 & 10.0 & 24.0 \\
 (3) & \textsc{Task Adapters (+bt)} & 42.2 & 35.9 & 33.5 & 30.9 & 39.2 & 25.5 & 33.5 & 23.6 & 22.2 & 18.3 & 13.2 &  28.9 \\
 & \textsc{Denois. Adapt. (+bt)} & 42.3 & 37.8 & \textbf{39.0} & \textbf{31.6} & \textbf{42.6} & \textbf{31.2} & \textbf{35.1} & \textbf{25.7} & \textbf{29.3} & 21.8 & \textbf{16.4} & \textbf{32.0} \\ 
 \midrule
 \midrule
\noalign{\smallskip} 
\multicolumn{14}{c}{$\bm{en}\to~\bm{zz}$} \\
\noalign{\smallskip} 
         & & es & nl & hr & uk & sv & lt & id & fi & et & ur & kk & \textsc{avg-11}  \\ \midrule
 (1) & \textsc{Bilingual} & \textbf{40.3} & \textbf{32.8} & \textbf{27.6} & \textbf{19.9} & \textbf{31.5} & 13.2 & 21.4 & 9.5 & 6.8 & 2.4 & 0.4 & \textbf{19.5} \\ \midrule
 & \textsc{Mbart-ft} & 1.3   & 1.9   & 1.8   & 0.8   & 1.6   & 0.9 & 1.6 & 1.4   & 0.7   & 0.6   & 0.4   & 1.3  \\
 (2) & \textsc{Task Adapters} & 2.0   & 2.0   & 2.1   & 1.0   & 1.5   & 0.9 & 0.8  & 1.6   & 1.1   & 0.9   & 0.5   & 1.4  \\
 & \textsc{Denois. Adapters} & 28.4  & 21.6  & 19.0  & 12.2  & 22.9  & 11.0  & 23.8 & 10.1  & 12.7 & 9.6 & 3.8 & 15.9 \\ \midrule
 & \textsc{Mbart-ft (+bt)} & 30.9 & 22.0 & 20.0 & 14.2 & 22.7 & 13.7 & 20.2 & 9.4 & 14.1 & 5.7 & 3.5 &  16.3 \\
 (3) & \textsc{Task Adapters (+bt)} & 31.5 & 22.4 & 21.9 & 15.7 & 25.3 & 14.6 & 22.9 & 10.1 & 15.2 & 9.4 & 4.2 & 17.6\\
 &\textsc{Denois. Adapt. (+bt)} & 32.2 & 22.9 & 23.1 & 15.4 & 27.1 & \textbf{16.3} & \textbf{24.4} & \textbf{11.7} & \textbf{17.1} & \textbf{11.7} & \textbf{4.9} & 18.9 \\ \bottomrule
 \end{tabular}
 \caption{Unsupervised translation to and from English. Only \textsc{Bilingual} is trained on parallel data and serves as reference. Block (2) is without back-translation, with only \textsc{Denois.\,Adapters} using monolingual data.
Block (3) uses the same amount of back-translation for all systems. Languages are presented by decreasing amount of parallel data used for training the bilingual baselines.}
 \label{tab:unsupervised-results}
 \end{table*}

\paragraph{Learning adapters from monolingual data}

We train the denoising adapters on a \textit{denoising} task, which aims to reconstruct text from a version corrupted with a noise function similar to mBART pretraining. Formally, we train denoising adapters \textbf{D} to \alex{minimize} $L_{D_{xx}}$:
\begin{eqnarray*}
L_{D_{xx}} = \sum_{T \in xx}-log P(T|g(T); \textbf{D}_{xx})
\end{eqnarray*}

\noindent where $T$ is a sentence in language $xx$ and $g$ is the noise function. We train denoising adapters on monolingual data for each language separately, including the unsupervised languages. This provides a high degree of flexibility for the later stages, such as unsupervised MT. During monolingual training, adapters are injected into layers of mBART, but only the adapter parameters are updated. The other parameters of the model stay frozen.
As noise function $g$, we use \textit{span} masking following mBART \cite{liu-etal-2020-multilingual} pretraining. A span of text with length $\ell$ (randomly sampled by a Poisson distribution) is replaced with the mask token. 

\paragraph{Multilingual MT fine-tuning with auxiliary parallel data}

After denoising adapters are trained for each language, 
the mBART model in which \textit{all} adapters are inserted is fine-tuned on the auxiliary multilingual English-centric parallel data. This step is required to force the model to learn how to use and combine denoising adapters for the translation task. 
During fine-tuning, we only update the parameters of the decoder's cross-attention, similarly to \citet{cooper-stickland-etal-2021-recipes} to limit the computational cost and mitigate catastrophic forgetting. 
The remaining parameters, including the newly plugged-in adapters are kept frozen at this stage.
When translating from language $xx$ to language $yy$, only the encoder denoising adapters $\textbf{D}_{xx}^E$ and decoder denoising adapters $\textbf{D}_{yy}^D$ are activated, as shown in Fig.~\ref{fig:nmt}. 

\paragraph{Multilingual UNMT process}

To summarize, we propose the following 2-stage training process for multilingual unsupervised MT: (1) Training denoising adapters within mBART, separately on each language's monolingual data; (2) Fine-tuning the cross-attention of a mBART augmented with the denoising adapters.

Fig.~\ref{fig:model} gives an overview of this process.
Our approach enables to use the final model for both supervised translation and unsupervised translation. 
For an unseen language \textit{zz} that has no parallel data, denoising adapters $\mathbb{D}^E_{zz}$ and $\mathbb{D}^D_{zz}$ can be trained on monolingual data and then combined with other existing languages for source/target side unsupervised translation. 
Denoising adapters not only allow us to skip back-translation, but also provide a high level of modularity and flexibility.
Except for the second step that uses only languages with parallel data, no additional joint training is needed.
As we show in Sect.~\ref{sec:new-languages}, by using denoising adapters, a new language which is not included in pretraining, can also be added successfully to mBART and used for unsupervised MT.
Note that all those new languages are however covered by the tokenizer (which is trained on 100 languages).

\section{Experimental Setup}
\label{sec:exp}

\paragraph{Dataset}

We use TED talks \cite{qi-etal-2018-pre} to create an English-centric (\textit{en}) multilingual dataset by picking 20 languages with different training size ranging from 214k (\textit{ar}) to 18k (\textit{hi}) parallel sentences. For multilingual UNMT evaluation, in addition to the 20 training languages, we select 17 ``unsupervised'' languages, 6 of which are unknown to mBART \cite{tang2020multilingual}. To train the denoising adapters, we use Wikipedia\footnote{We used the latest Wikipedia dumps as of \textit{15.02.2021}} \ahmet{and News Crawl\footnote{http://data.statmt.org/news-crawl/}} with maximum 20M sentences per language. Details of languages and training datasets are given in Appendix \ref{app:langs}

\paragraph{Baselines}

We compare our approach with the following baselines: 
(1) \textsc{Bilingual}, baseline bilingual models trained on TED talks. These are small Transformer models trained separately on each language direction, using the same settings as \citet{philip-etal-2020-monolingual}. Note that these models do not have any pretraining and they are trained from scratch. (2) \textsc{Mbart-ft}, standard fine-tuning of mBART \cite{liu-etal-2020-multilingual} on the multilingual MT task. (3) \textsc{Task\,Adapters}, multilingual fine-tuning for \textit{language-agnostic} MT adapters and cross-attention on top of mBART, similarly to \citet{cooper-stickland-etal-2021-recipes}.

The bilingual models and all the mBART variants are fine-tuned on the same English-centric multilingual parallel data. 

\begin{table*}[ht]
\small
\centering
\def\arraystretch{1.15}
\setlength\tabcolsep{5.5pt}
\begin{tabular}{@{}lrrrrrrrcrrrrrrr@{}}
\multicolumn{1}{c}{} & \multicolumn{7}{c}{$\bm{zz}\to~\bm{en}$} & \multicolumn{8}{c}{$\bm{en}\to~\bm{zz}$} \\ 
\cline{2-8} \cline{10-16} \noalign{\smallskip} 
& bg & hu & sr & el & da & be & \textsc{avg-6} &  & bg & hu & sr & el & da & be & \textsc{avg-6} \\
\textsc{Bilingual} & \textbf{40.7} & 27.3 & 34.2 & \textbf{38.7}  & 41.1 & 3.12 & 30.9 & & \textbf{35.1}  & \textbf{19.2}  & \textbf{21.3} & \textbf{32.2} & \textbf{36.4} & 2.14 & \textbf{24.4} \\ \cline{1-8} \cline{10-16} \noalign{\smallskip} 
\textsc{Mbart-ft} & 8.8  & 1.0  & 18.9 & 0.2 & 5.2 & 2.8 & 6.2 & & -  & -  & -   & -  & -  & - & - \\
\textsc{Task A.} & 11.9  & 1.3  & 24.8  & 0.5 & 8.3 & 4.6  & 8.6 & & -  & -  & -   & -  & -  & - & - \\
\textsc{Denois. A.} & 39.8  & \textbf{27.5}  & \textbf{36.9}  & 34.6  & \textbf{45.5}  & \textbf{28.4}  & \textbf{35.5} & & 24.1  & 11.1  & 8.6   & 16.1  & 25.7  & \textbf{12.1}  & 16.3 \\
\hline
\end{tabular}
\caption{Unsupervised translation performance for languages that are new to mBART.}
\label{tab:new-languages}
\end{table*}

\paragraph{Multilingual MT training details} We train mBART-based models by using a maximum batch size of 4k tokens and accumulated gradients over 5 update steps with mixed precision \cite{ott-etal-2018-scaling} for 120k update steps. We apply Adam \cite{kingma2014adam} with a polynomial learning rate decay, and a linear warmup of \numprint{4000} steps for a maximum learning rate of $0.0001$. Additionally, we use dropout with a rate of $0.3$ and label smoothing with a rate of $0.2$. For efficient training, we filter out the unused tokens from the mBART vocabulary after tokenization of the training corpora (including both TED talks and monolingual datasets) which results a shared vocabulary of 210k tokens. Finally, following \citet{arivazhagan}, we use temperature-based sampling with $T=5$ to balance language pairs during training. As for bilingual baselines, we train these models for 25k updates on the TED talks bilingual data, with maximum 4k tokens per batch and accumulated gradients over 4 updates. Joint BPE models of size 8k are used for these models. All experiments are performed with the fairseq \cite{ott-etal-2019-fairseq} library.

\paragraph{Adapter Modules} We used the  architecture of \citet{philip-etal-2020-monolingual} for the adapters with a bottleneck dimension of 1024 in all experiments. 
As noising function for our denoising adapters, we mask 30\% of the words in each sentence with a span length that is randomly sampled by a Poisson distribution ($\lambda=3.5$) as same with mBART \cite{liu-etal-2020-multilingual}. We train these adapters separately for each language for 100k training steps by using a maximum batch size of 4k tokens, accumulating gradients over 8 update steps and a maximum learning rate of $0.0002$. Other hyperparameters are the same as in the NMT training. 

\paragraph{Back-translation}

As second part of the evaluation, we also used \textit{offline} back-translation for (1) comparing \deo{} with baselines that are additionally trained on back-translated synthetic parallel data; 
and (2) measuring the impact of back-translation when it is applied in conjunction with denoising adapters. Following \citet{garcia2020harnessing} ---that shows the effectiveness of offline back-translation for multilingual UNMT---,
we back-translate the monolingual data into English (\textit{en}) for each unsupervised language \textit{zz} with the respective model. After that, we fine-tune the corresponding model by using its back-translated parallel data in a single (bilingual) direction for both \textit{zz}$\to$\textit{en} and \textit{en}$\to$\textit{zz} separately. For fine-tuning we either fine-tune the full model (\textsc{Mbart-ft}) or only update adapters' and cross-attention's parameters (\textsc{Task A.}, \textsc{Denoising A.}) for 120k additional steps. For fair comparison, we limit the monolingual data to 5M for both denoising adapter training and back-translation in these experiments. Note that this procedure is both memory and time-intensive operation as it requires back-translating a large amount of monolingual data, and it also results in an extra bilingual model to be trained for each unsupervised language and for all models that are evaluated. 

\section{Results}
\label{sec:main-results}

Table~\ref{tab:unsupervised-results} shows translation results for 11 languages that have no parallel data, in \textit{zz}$\to$\textit{en} and \textit{en}$\to$\textit{zz} directions. 
The first two blocks in each direction, (1) and (2), give unsupervised translation results without using back-translation. 

For \textit{zz}$\to$\textit{en}, the two baselines \mbart and \task are quite decent: the ability of mBART to encode the unsupervised source languages and its transfer to NMT using auxiliary parallel data provide good multilingual unsupervised NMT performance. Among the two baselines, task-specific MT adapters better mitigate catastrophic forgetting, ensuring the model does not overfit to the supervised languages and to benefit more from multilingual fine-tuning which results in +5.4 BLEU compared to standard fine-tuning. 
Our approach, however, outperforms the two mBART baselines and the bilingual models: denoising adapters are superior for all languages compared to \mbart and \task and result in respectively +8.6 and +3.2 BLEU on average.
Finally, it even performs better than the \textit{supervised} bilingual models for most languages (all but \textit{es} and \textit{nl}). 

\begin{figure}[t]
    \includegraphics[scale=0.53]{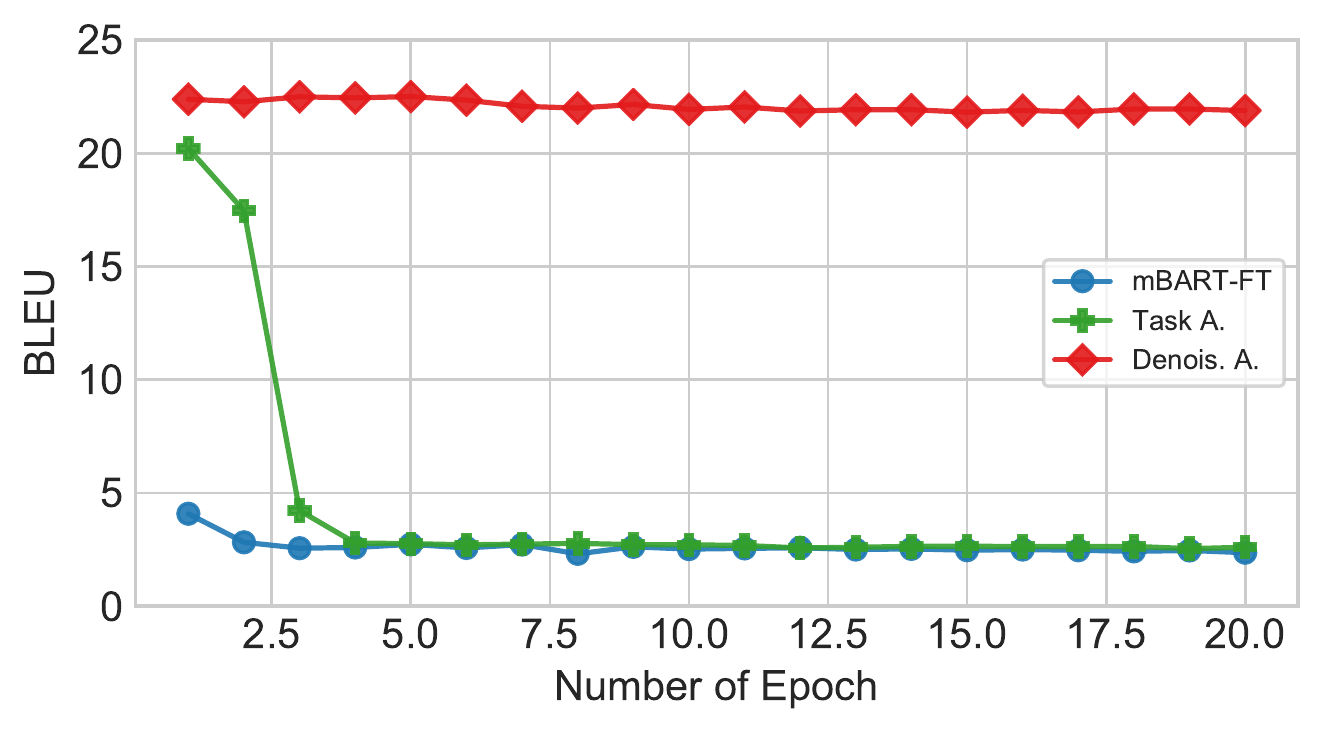}
    \centering
  \caption{\textit{en}$\to$\textit{nl}~(unsupervised) performance on validation data during mBART fine-tuning for each model.}
  \label{fig:valid.en-nl}
\end{figure}

For the \textit{en}$\to$\textit{zz} direction, \laurent{the two baselines \mbart and \task are ineffective}, showing the limitation of mBART pretraining for \laurent{multilingual UNMT}
when translating \textit{from} English. 
A possible explanation for this is the fact that these models have learnt to encode English with only auxiliary target languages; and the transfer from mBART to NMT has made the decoder forget how to generate text in the 11 unsupervised languages we are interested in. Fig.~\ref{fig:valid.en-nl} shows unsupervised translation performance for \textit{en}$\to$\textit{nl} in validation set during mBART fine-tuning. As opposed to our approach, the low start in \mbart and the quick drop in \task confirm the forgetting in generation. 
However, denoising adapters that leverage monolingual training for language-specific representations enable the final model to achieve high translation quality without any parallel data even without back-translation. 
Denoising adapters also outperform the supervised bilingual models trained with less than 50k parallel sentences.

\begin{table*}[ht]
\small
\centering
\begin{tabular}{@{}lrrrrrrrrrrrr@{\hskip 0.25in}c@{}}
\toprule
\multicolumn{14}{c}{$\bm{xx}\to~\bm{en}$} \\
\noalign{\smallskip} 
& ar & he & ru & it & fr & tr & pl & vi & de & fa & cs & hi & \textsc{avg-20}  \\ \midrule
\textsc{Bilingual} & 33.0  & 39.0  & 26.0  & 39.7  & 41.7  & 27.7  & 25.5  & 28.3  & 37.4  & 28.9  & 28.7  & 9.7   & 28.0 \\ \midrule
\textsc{Mbart-ft} & \textbf{35.2}  & 40.4  & \textbf{29.2}  & 42.5  & 44.2  & 31.1  & 29.0  & 31.2  & 40.9  & \textbf{33.7}  & 34.7  & \textbf{31.6}  & \textbf{33.0} \\
\textsc{Task Adapters}   & 33.5  & 38.9  & 28.8  & 41.9  & 43.9  & 30.4  & 28.4  & 31.1  & 40.3  & 32.3  & 34.4  & 30.6  & 32.3 \\
\textsc{Lang. Adapters} & 35.2  & \textbf{40.5}  & 29.1  & \textbf{42.6}  & \textbf{44.4}  & \textbf{31.3}  & \textbf{29.1}  & \textbf{31.5}  & \textbf{41.3}  & 33.3  & \textbf{35.0}  & 30.0  & 32.9 \\
\textsc{Denois. Adapters}     & 32.6  & 38.0  & 27.6  & 41.0  & 42.9  & 28.8  & 27.6  & 29.9  & 39.3  & 31.2  & 34.0  & 27.1  & 30.5 \\
\midrule
\midrule
\noalign{\smallskip} 
\multicolumn{14}{c}{$\bm{en}\to~\bm{xx}$} \\
\noalign{\smallskip} 
& ar   & he   & ru   & it   & fr   & tr   & pl   & vi   & de   & fa   & cs   & hi   & \textsc{avg-20}  \\ \midrule
\textsc{Bilingual} & \textbf{17.2} & \textbf{27.5} & 20.5 & 35.4 & 40.7 & 16.5 & 18.2 & 29.4 & 30.0 & 15.0 & 20.8 & 10.7 & 22.4 \\ \midrule
\textsc{Mbart-ft}  & 16.6 & 25.8 & \textbf{21.6} & \textbf{36.8} & \textbf{41.6} & \textbf{18.2} & \textbf{19.1} & \textbf{31.2} & \textbf{31.8} & \textbf{16.8} & \textbf{23.3} & 22.2 & \textbf{24.5} \\
\textsc{Task Adapters}  & 15.6 & 24.3 & 21.1 & 35.8 & 41.0 & 17.6 & 18.2 & 30.4 & 31.0 & 16.4 & 22.4 & \textbf{22.3} & 23.8 \\
\textsc{Lang. Adapters}   & 16.0 & 24.9 & 21.1 & 36.0 & 41.2 & 17.5 & 18.8 & 30.8 & 31.2 & 16.6 & 22.7 & 21.4 & 24.0 \\
\textsc{Denois. Adapters}     & 14.4 & 21.5 & 19.5 & 33.1 & 38.8 & 15.8 & 17.3 & 29.5 & 28.9 & 15.3 & 21.2 & 17.8 & 21.7 \\
\bottomrule
\end{tabular}
\caption{Supervised translation results to and from English for auxiliary languages. Languages are presented by decreasing amount of parallel data used for training the bilingual baselines.
Due to lack of space we only show individual results on 12 representative languages. Full list of results are given in Appendix~\ref{app:full-results}}.
\label{tab:supervised-results}
\end{table*}

\paragraph{Impact of back-translation}

3rd blocks (3) in Table~\ref{tab:unsupervised-results} show the unsupervised translation results after models are fine-tuned with offline back-translated parallel data. Note that in this step each model is fine-tuned for a single language-pair and only one direction. 

For \textit{zz}$\to$\textit{en}, although back-translation slightly improves the results, the overall impact of  back-translation is very limited for all models including our approach. Interestingly, for \textit{ur} the back-translation decreased the performance. We relate this to the domain difference between test (TED talks) and back-translated data (Wikipedia/News). Here, denoising adapters without back-translation still provide superior unsupervised translation quality compared to baselines even after the back-translation. 

For \textit{en}$\to$\textit{zz}, the back-translation significantly increased translation results: +15.0, +16.2 and +3.0 BLEU for \textsc{Mbart-ft}, \task and \textsc{Denoising Adapters} respectively. 
We hypothesize that the huge boost in the baselines scores is due to the fact that training on the back-translated parallel data allows these models to recover generation ability in the target languages. However, our approach outperforms baselines in all languages, showing that denoising adapters can be used jointly with back-translation for further improvements. Finally, denoising adapters without back-translation (2) are still competitive with the mBART baselines. 

\label{sec:backtranslation}

\begin{figure}[t]
  \begin{subfigure}[b]{0.49\textwidth}
    \includegraphics[width=\textwidth]{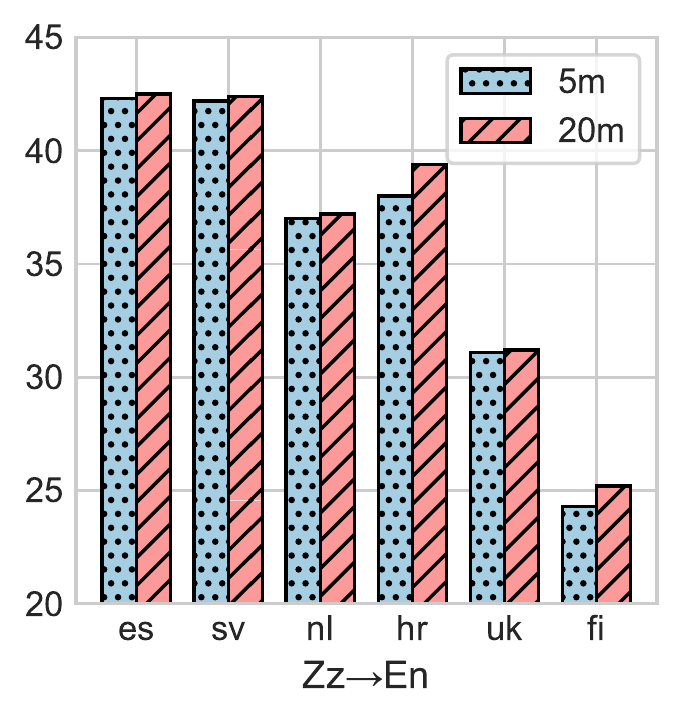}
    \label{fig:datasize-en}
    \centering
  \end{subfigure} 
  \begin{subfigure}[b]{0.49\textwidth}
    \includegraphics[width=\textwidth]{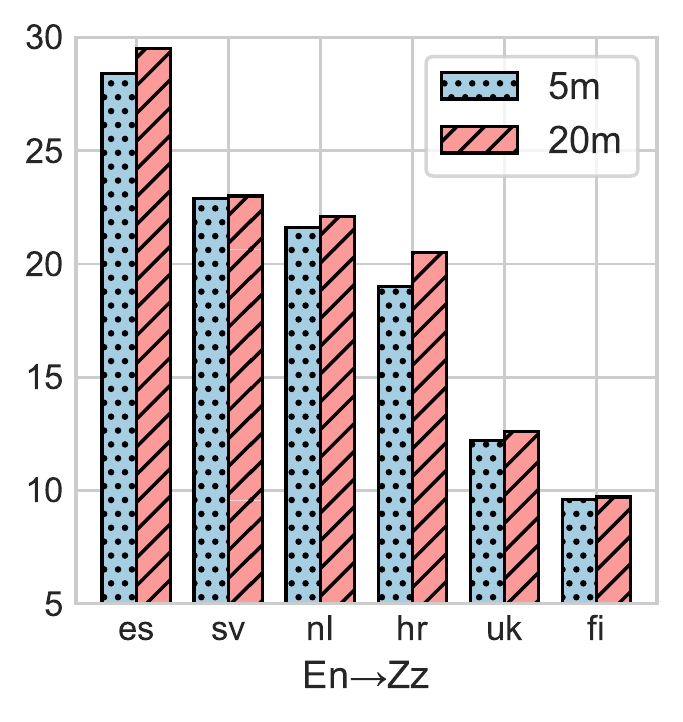}
    \label{fig:datasize-xx}
  \end{subfigure}
  \vspace{-1cm}
  \caption{Unsupervised translation results (BLEU) for denoising adapters trained on 5m and 20m sentences.}
  \label{fig:datasize}
\end{figure}

\section{Analysis and Discussion}

\subsection{Denoising adapters for languages unknown to mBART}
\label{sec:new-languages}

All the languages considered so far (in Table~\ref{tab:unsupervised-results}) were included in the mBART-50 pretraining data \cite{tang2020multilingual}. 
Here, we also evaluate our model on languages that are new to mBART-50,\footnote{Note that mBART uses the same sentencepiece vocabulary \cite{kudo-richardson-2018-sentencepiece} as XLM-R \cite{conneau2019unsupervised} which is trained on 100 languages including the ones we used for evaluation.} to test whether our denoising adapters can be used to extend the translation model incrementally to new languages using monolingual data.
After training our denoising adapters, we insert them into the existing NMT model of Sect.~\ref{sec:model} for unsupervised MT with no additional NMT training. 
Denoising adapter layers are trained the same way as before with only a small difference: we update the output projection layer of mBART together with adapter layers to improve language-specific decoding.

Table~\ref{tab:new-languages} shows the results in both directions 
for the bilingual baselines and other mBART variants that are fine-tuned with only auxiliary parallel data. 
For \textit{zz}$\to$\textit{en} although the models are trained on English-centric multilingual parallel corpora with related languages, mBART baselines still have very poor unsupervised MT performance. 
Denoising adapters, however, with the advantage of monolingual data and modular training, display competitive or better results even compared to supervised bilingual baselines. 
Moreover, for the \textit{en}$\to$\textit{zz} direction, it provides a reasonable level of unsupervised translation quality that can be used with back-translation for further improvements. 
Note that, since neither mBART pretraining nor the multilingual fine-tuning include those new languages, the other baselines are not able to translate in these directions.

Overall these results confirm that denoising adapters offer an efficient way to extend mBART to new languages. Moreover, taken together with the other results (Sect.~\ref{sec:main-results}), unsupervised translation quality for the missing languages without additional NMT training demonstrates the 
effectiveness of our approach.

\subsection{Monolingual data size}
\label{sec:monolingual}

To see the impact of the monolingual data size that is used for training of denoising adapters, we additionally trained adapters on larger data for 6 languages (\textit{es, sv, nl, hr, uk, fi}). Fig.~\ref{fig:datasize} shows the unsupervised translation results when they are trained on two different data sizes: 5m and 20m sentences. Interestingly, for a majority of languages, the performance improvement is very limited with increase in data size. This confirms that denoising adapters achieve competitive performance without the need of a huge amount of monolingual data. 

\begin{table*}[t]
\small
\centering
\begin{tabular}{@{}lcccc@{\hskip 0.3in}cccc@{}}
\toprule
FLoRes \textit{devtest} & \multicolumn{4}{c}{ne} & \multicolumn{4}{c}{si} \\ \midrule
& \textsc{Bleu} & \textsc{chrF} & \textsc{Comet} & \textsc{BertScore} & \textsc{Bleu} & \textsc{chrF} & \textsc{Comet} & \textsc{BertScore} \\ 
\noalign{\smallskip}
\textsc{Mbart} \cite{liu-etal-2020-multilingual} & 17.9* & - & - & - & 8.1* & - & - & -\\
\midrule
\textsc{Mbart-ft}  & 15.7 & 42.6 & 19.6 & 49.1 & 7.6 & 32.4 & -6.0 & 37.4 \\
\textsc{Denois.\,Adapt.}  & 18.1 & 44.0 & 31.5 & 54.8 & 11.4 & 37.0 & 15.0 & 46.3 \\
\bottomrule
\end{tabular}
\caption{Unsupervised translation results on the FLoRes \textit{devtest} sets~\cite{guzman2019flores}. \textsc{Mbart-ft} and \textsc{Denois.\,Adapt.} are trained only on~ \textbf{\textit{hi}}$\to$\textbf{\textit{en}}.
Note that we used mBART-50 for our replication of \textsc{Mbart-ft} and \textsc{Denoising\,Adapters}, however the original paper results are based on mBART-25.
\textsc{Mbart}~(*) results are taken from the paper \cite{liu-etal-2020-multilingual} and are the only evaluation results in this paper not done by ourselves.
}
\label{tab:sota}
\end{table*}

\subsection{Supervised translation}
\label{sec:m2m-translation}

Finally, we evaluate the baselines and our model on the supervised languages (i.e. the \emph{auxiliary} languages with access to parallel data). Table~\ref{tab:supervised-results} shows BLEU scores for \textit{xx}$\to$\textit{en} and \textit{en}$\to$\textit{xx} directions. In this setting, in addition to the main baselines, we include \textsc{Language Adapters}~\citep{philip-etal-2020-monolingual}, which correspond to fine-tuning both \textit{language-specific} MT adapters and cross-attention on top of mBART only with parallel data.
As expected, for both directions multilingual fine-tuning of mBART (\textsc{Mbart-ft}) performs the best on average.
The performance of \textsc{Lang.\,Adapters} is on par with full fine-tuning. 
For \textit{xx}$\to$\textit{en}, it outperforms full fine-tuning in 10 out of 20 language pairs, with the a very similar overall score.
For \textit{en}$\to$\textit{xx}, it has only -0.5 BLEU on average.
\textsc{Task adapters} have slightly lower translation performance than these other two models on both directions. 
Nonetheless, on \textit{en}$\to$\textit{xx} direction, as the amount of parallel data decreases (see Sect.~\ref{app:langs}), the gap between this model and full \mbart reduces, confirming that task adapters are beneficial for small data and distant language pair conditions \cite{cooper-stickland-etal-2021-recipes}.
As for multilingual fine-tuning with \textsc{Denois.\,Adapters}, although it has lower scores than other mBART variants, it still performs competitively with the bilingual baselines. 
It outperforms the bilingual baselines in \textit{xx}$\to$\textit{en} and gets -0.7 BLEU on average in \textit{en}$\to$\textit{xx}. 
Unlike other mBART variants, fine-tuning only the decoder's cross-attention seems to penalize performance.
Considering that denoising adapters are designed specifically for multilingual unsupervised MT, these results show that our approach still performs on a competitive level in the large-scale supervised multilingual NMT setup. 

\subsection{Comparison with state-of-the-art}
\label{sec:sota}

\ahmet{With the goal of providing a comparison point with a previously reported set-up that does not include back-translation, we replicate the \textit{language-transfer} results  reported in \cite[mBART]{liu-etal-2020-multilingual}.
For that, we fine-tune mBART-50 \cite{tang2020multilingual} on Hindi-English (\textit{hi}$\to$\textit{en})  parallel data from IITB \cite{kunchukuttan2017iit} and test the resulting model on two unseen languages, Nepali (\textit{ne}) and Sinhalese (\textit{si}), from the FLoRes dataset \cite{guzman2019flores} without any further training on back-translated data.
For \textsc{Denoising Adapters}, we trained adapters on monolingual data provided by FLoRes for all 4 languages (\textit{en, hi, ne, si}). 
Finally for MT transfer, we inserted these language-specific adapters to mBART, and updated cross-attention layers as in the previous experiments. 
Results are shown in Table \ref{tab:sota}.}

\ahmet{We compare results in terms of BLEU,\footnote{SacreBLEU~\cite{post-2018-call} signature:\\ \texttt{BLEU+c.mixed+\#.1+s.exp+tok.13a+v.1.5.0}} chrF \cite{popovic-2015-chrf}, {\sc Comet} \cite{rei-etal-2020-comet}\footnote{{\sc Comet} model: \texttt{wmt20-comet-da}} 
and BERT Score \cite{bert-score}.\footnote{Bert score hash code: \\ \texttt{roberta-large\_L17\_no-idf\_version=0.3.10\\(hug\_trans=4.10.0)-rescaled\_fast-tokenizer}} In all three metrics \textsc{Denoising\,Adapters} significantly outperform \textsc{Mbart-ft}, showing the effectiveness of denoising adapters for low resource languages, compared to a strong baseline. Note that since we used mBART-50 in our experiments, results for \textsc{Mbart-ft} are slightly different from the ones in original paper (mBART-25).}

\section{Conclusion}
\label{sec:conclusion}

We have presented denoising adapters, adapter modules trained on monolingual data with a \textit{denoising} objective, and a 2-step approach to adapt mBART by using these adapters for multilingual unsupervised NMT. Our experiments conducted on a large number of languages show that denoising adapters are very effective for unsupervised translation even without the need of  back-translation. Moreover, denoising adapters are complementary with back-translation; using them jointly improves the translation quality even further. We have also demonstrated that for a language new to mBART, denoising adapters offer an efficient way to extend mBART incrementally. Finally, although it is designed for unsupervised NMT, our approach still reaches competitive performance in supervised translation in a multilingual NMT setup. 

\ahmet{For the future direction, translating between two unseen languages may be considered as a natural extension of our work. As preliminary experiment, we addressed a language pair including two languages of the unsupervised setup: Spanish (\textit{es}) and Dutch (\textit{nl}). We inserted denoising adapters of those languages to encoder/decoder and directly used this model without further training for \textit{nl}$\to$\textit{es} and \textit{es}$\to$\textit{nl}. Although our auxilliary language pairs with parallel data are English-centric, these two models perform at a decent level (15.4, 7.2 BLEU respectively) and they could be a good starting point for further improvements.
Another  direction is to apply denoising adapters to domain adaptation, a use-case where back-translation is a standard solution to leverage monolingual data. 
}

\section*{Acknowledgements}
We would like to thank Vassilina Nikoulina, Asa Cooper Stickland, and Naver Lab Europe NLP team for discussions during the project. Furthermore, we thank Arianna Bisazza, Gosse Bouma, Gertjan van Noord and the anonymous reviewers for their feedback.

\newpage

\bibliography{emnlp2020}
\bibliographystyle{acl_natbib}

\appendix

\section{Appendix}

\subsection{Language Details}
\label{app:langs}


We build our experimental setup based on TED talks \cite{qi-etal-2018-pre}. Together with English (\textit{en}) as the center language, we choose 19 training languages. As unsupervised languages, we pick 17 languages without using their parallel data. For the language selection, we consider following criteria: 

\begin{itemize}
    \item Varying parallel data sizes; from \textit{en}$\leftrightarrow$\textit{ar} (214k) to \textit{en}$\leftrightarrow$\textit{hi} (18k) 
    \vspace{-2.5mm}
    \item Diversity in terms of language families. For unsupervised languages, we both select languages having close relation with training cluster (e.g. \textit{es}) and distant languages (e.g. \textit{fi}).
    \vspace{-2.5mm}
    \item Different monolingual data sizes: from 20M sentences (\textit{en}) to 900k sentences (\textit{ur}).
    \vspace{-2.5mm}
    \item The language list of mBART-50 \cite{tang2020multilingual}. Among 17 unsupervised languages, 11 are present and the remaining 6 languages are not included in the pretraining. Note that the mBART vocabulary consists of 100 languages that covers all these 17 languages. 
\end{itemize}

Details of languages are given in Table~\ref{tab:languages}. 
We report the amount of parallel data for all languages, including those where this is not used as it constitutes the training data for the supervised bilingual baselines.

\begin{table}[H]
\small
\begin{tabular}{@{}lllrr@{}}
\toprule
\multirow{2}{*}{Language}&\multirow{2}{*}{Code}&Language & Mono. & Parallel \\ 
& & family & data (M) & data (k) \\ \midrule
English    & en   & Germanic       & 20              & 250          \\
Arabic     & ar   & Semitic        & 20              & 214          \\
Hebrew     & he   & Semitic        & 6.9             & 211          \\
Russian    & ru   & Slavic         & 20              & 208          \\
Korean     & ko   & Korean         & 17              & 205          \\
Italian    & it   & Romance        & 20              & 204          \\
Japanese   & ja   & Japonese       & 20               & 204          \\
Chinese    & zh   & Sino-Tibetan   & 18             & 199          \\
French     & fr   & Romance        & 20              & 196          \\
Portuguese & pt   & Romance        & 20              & 192          \\
Turkish    & tr   & Turkic         & 19              & 182          \\
Romanian   & ro   & Romance        & 20              & 180          \\
Polish     & pl   & Slavic         & 17              & 176          \\
Vietnamese & vi   & Austri-Asiatic & 6.7             & 171          \\
German     & de   & Germanic       & 20              & 167          \\
Persian    & fa   & Iranian        & 5.7             & 150          \\
Czech      & cs   & Slavic         & 20              & 103          \\
Thai       & th   & Tai-Kadai      & 2.2             & 98           \\
Burmese    & my   & Sino-Tibetan   & 0.2             & 21           \\
Hindi      & hi   & Indic          & 20              & 18           \\
\noalign{\smallskip} 
\hdashline 
\noalign{\smallskip} 
Spanish    & es   & Romance        & 5              &  \hl{196}    \\
Dutch      & nl   & Germanic       & 5              &  \hl{183}   \\
Crotian    & hr   & Slavic         & 5             &  \hl{122}   \\
Ukrainian  & uk   & Slavic         & 5              &  \hl{108}    \\
Indonesian & id   & Austronesian   & 4.8             &  \hl{~~87}   \\
Swedish    & sv   & Germanic       & 5              &   \hl{~~56}    \\
Lithuanian & lt   & Slavic         & 4.5             &  \hl{~~41}    \\
Finnish    & fi   & Finnic         & 5              &   \hl{~~24}   \\
Estonian   & et   & Finnic         & 5              &  \hl{~~10}  \\
Urdu       & ur   & Indic          & 0.9             & \hl{~5.9}  \\
Kazakh     & kk   & Turkic         & 3.4             &   \hl{~3.3}\\
\noalign{\smallskip} 
\hdashline 
\noalign{\smallskip} 
Bulgarian  & bg   & Slavic         & 20              &  \hl{174}  \\
Hungarian  & hu   & Uralic         & 20              &   \hl{147}  \\
Serbian    & sr   & Slavic         & 8.7             & \hl{136} \\
Greek      & el   & Greek          & 11              &   \hl{134}    \\
Danish     & da   & Germanic       & 2.9             &\hl{~~44}    \\
Belarusian & be   & Slavic         & 1.7             &\hl{~4.5}    \\ \bottomrule
\end{tabular}
\caption{Languages that are used in the experiments. The first block shows training languages with parallel data, the second block refers unsupervised languages that are included in mBART-50 \cite{tang2020multilingual} and the last block gives languages new to mBART-50. \hl{Greyed out} numbers indicate data that is only used for the supervised bilingual baselines.}
\label{tab:languages}
\end{table}

\subsection{Experimental Details}
\label{app:imp}

\begin{table}[H]
\centering
\small
\begin{tabular}{@{}ll@{}}
\toprule
Hyper-Parameter & Value \\ \midrule
Architecture & mbart\_large \\
Optimizer                  & Adam     \\
$\beta_1,~\beta_2$         & 0.9, 0.98 \\
Weight decay               & 0.01     \\
Label smoothing            & 0.2    \\
Dropout                    & 0.3      \\
Attention dropout          & 0.1      \\
Batch size   & 4k (tokens)      \\ 
Update frequency   & 5      \\
Warmup updates   & 4000      \\
Total number of updates  & 120k       \\
Max learning rate         & 0.0001 \\
Learning rate scheduler  & polynomial\_decay \\
Temperature (sampling) & 5 \\
\noalign{\smallskip} 
\hdashline 
\noalign{\smallskip}
Adapter dim. & 1024 \\
Noise function & span\_masking \\
Mask ratio & 0.3 \\
Mask random replace ratio & 0.1 \\
Poisson lambda & 3.5 \\
Update frequency   & 8      \\
Total number of updates  & 100k       \\
Max learning rate         & 0.0002 \\
\bottomrule
\end{tabular}
\caption{Fairseq hyperparameters for our experiments. The first block gives the base settings used for \textsc{Mbart-ft} and the second block provides the details for the \textsc{Denoising A.} when it differs from the base settings.} 
\label{tab:hyperparams}
\end{table}

We use the fairseq library \cite{ott-etal-2019-fairseq} to conduct our experiments. The hyperparameters used for fairseq are given in Table~\ref{tab:hyperparams}. For the parallel data, we used the TED talks corpus without any other pre-processing than the mBART SentencePiece tokenization. For the monolingual data, we downloaded the Wikipedia articles together with News Crawl datasets for each language. For Wikipedia articles, we pre-processed the data by using WikiExtractor \cite{Wikiextractor2015} and tokenized sentences\footnote{We use \url{https://github.com/microsoft/BlingFire} for basic tokenization.}. We train denoising adapters and fine-tune mBART models by using 4 Tesla V100 GPUs with mixed precision. 
Finally, for evaluation over the TED talks test sets, we used SacreBLEU \cite{post-2018-call} \footnote{\texttt{BLEU+c.mixed+\#.1+s.exp+tok.none+v.1.5.0}. For Chinese and Japanese we use \texttt{--language-pair} option for language specific tokenization}. The best checkpoint is chosen according to validation BLEU scores for NMT models and for denoising adapters we use the last checkpoint for each language. 

\subsection{Full List of Supervised Translation Results}
\label{app:full-results}

\begin{table}[H]
\small
\centering
\begin{tabular}{@{}llllll@{}}
\toprule
\noalign{\smallskip}
& \rotatebox[origin=c]{65}{\textsc{Bilingual}} &\rotatebox[origin=c]{65}{\textsc{Mbart-ft}} & \rotatebox[origin=c]{65}{\textsc{Task A.}} & \rotatebox[origin=c]{65}{\textsc{Lang. A.}} & \rotatebox[origin=c]{65}{\textsc{Denois. A.}} \\ 
\noalign{\smallskip}
\noalign{\smallskip}
\multicolumn{6}{c}{$\bm{xx}\to~\bm{en}$} \\
\noalign{\smallskip}
ar     & 33.0 & 35.2 & 33.5 & 35.2 & 32.6 \\
he     & 39.0 & 40.4 & 38.9 & 40.5 & 38.0 \\
ru     & 26.0 & 29.2 & 28.8 & 29.1 & 27.6 \\
ko     & 20.4 & 23.3 & 22.6 & 22.8 & 20.9 \\
it     & 39.7 & 42.5 & 41.9 & 42.6 & 41.0 \\
ja     & 14.9 & 17.3 & 17.1 & 17.4 & 15.3 \\
zh     & 21.2 & 24.1 & 23.2 & 23.7 & 22.1 \\
fr     & 41.7 & 44.2 & 43.9 & 44.4 & 42.9 \\
pt & 46.2 & 48.7 & 48.1 & 49.2 & 47.6 \\
tr     & 27.7 & 31.1 & 30.4 & 31.3 & 28.8 \\
ro     & 36.5 & 40.3 & 39.6 & 40.1 & 39.0 \\
pl     & 25.5 & 29.0 & 28.4 & 29.1 & 27.6 \\
vi     & 28.3 & 31.2 & 31.1 & 31.5 & 29.9 \\
de     & 37.4 & 40.9 & 40.3 & 41.3 & 39.3 \\
fa     & 28.9 & 33.7 & 32.3 & 33.3 & 31.2 \\
cs     & 28.7 & 34.7 & 34.4 & 35.0 & 34.0 \\
th     & 22.1 & 28.0 & 26.8 & 27.9 & 21.9 \\
my     & 5.2  & 21.8 & 20.8 & 21.0 & 12.1 \\
hi     & 9.7  & 31.6 & 30.6 & 30.0 & 27.1 \\
avg    & 28.0 & 33.0 & 32.3 & 32.9 & 30.5 \\
\midrule
\midrule
\noalign{\smallskip} 
\multicolumn{6}{c}{$\bm{en}\to~\bm{xx}$} \\
\noalign{\smallskip} 
ar     & 17.2 & 16.6 & 15.6 & 16.0 & 14.4 \\
he     & 27.5 & 25.8 & 24.3 & 24.9 & 21.5 \\
ru     & 20.5 & 21.6 & 21.1 & 21.1 & 19.5 \\
ko     & 8.4  & 9.1  & 8.5  & 8.9  & 7.6  \\
it     & 35.4 & 36.8 & 35.8 & 36.0 & 33.1 \\
ja     & 13.4  & 15.6 & 14.6 & 15.5 & 13.0 \\
zh     & 24.1  & 22.4 & 21.5 & 22.5 & 20.0 \\
fr     & 40.7 & 41.6 & 41.0 & 41.2 & 38.8 \\
pt & 40.5 & 41.2 & 40.2 & 40.7 & 38.5 \\
tr     & 16.5 & 18.2 & 17.6 & 17.5 & 15.8 \\
ro     & 27.2 & 28.5 & 27.9 & 28.0 & 25.8 \\
pl     & 18.2 & 19.1 & 18.2 & 18.8 & 17.3 \\
vi     & 29.4 & 31.2 & 30.4 & 30.8 & 29.5 \\
de     & 30.0 & 31.8 & 31.0 & 31.2 & 28.9 \\
fa     & 15.0 & 16.8 & 16.4 & 16.6 & 15.3 \\
cs     & 20.8 & 23.3 & 22.4 & 22.7 & 21.2 \\
th     & 18.8 & 19.9 & 19.6 & 19.4 & 16.0 \\
my     & 11.8 & 24.7 & 24.8 & 23.6 & 18.0 \\
hi     & 10.7 & 22.2 & 22.3 & 21.4 & 17.8 \\
avg    & 22.4 & 24.5 & 23.8 & 24.0 & 21.7 \\
\bottomrule
\end{tabular}
\caption{Full list of supervised translation results to and from English for auxiliary languages. Languages are presented by decreasing amount of parallel data used for training the bilingual baselines.}
\end{table}

\end{document}